\pdfoutput=1
\documentclass{article}

\usepackage[nonatbib, final]{neurips_2021_preregistration}

\usepackage{graphicx}
\usepackage[abbreviations]{glossaries-extra}

\glssetcategoryattribute{acronym}{indexonlyfirst}{true}

\newabbreviation{mlp}{MLP}{multilayer perceptron}
\newabbreviation{cbmlp}{CbMLP}{coordinate-based \gls{mlp}}
\newabbreviation{nerf}{NeRF}{Neural Radiance Field}
\newabbreviation{dct}{DCT}{discrete cosine transform}

\usepackage[utf8]{inputenc} 
\usepackage[T1]{fontenc}    
\usepackage[hidelinks]{hyperref}       
\usepackage{url}            
\usepackage{booktabs}       
\usepackage{amsfonts}       
\usepackage{nicefrac}       
\usepackage{microtype}      
\usepackage{xcolor}         

\title{Neural Weight Step Video Compression}

\author{%
  Mikolaj Czerkawski \And Javier Cardona \And Robert Atkinson \And Craig Michie \And Ivan Andonovic \And Carmine Clemente \And Christos Tachtatzis \And\\
  University of Strathclyde\\
99 George St, G1 1RD\\Glasgow, United Kingdom\\
  \texttt{mikolaj.czerkawski@strath.ac.uk} \\
}

\begin{document}

\maketitle

\begin{abstract}
    A variety of compression methods based on encoding images as weights of a neural network have been recently proposed. Yet, the potential of similar approaches for video compression remains unexplored. In this work, we suggest a set of experiments for testing the feasibility of compressing video using two architectural paradigms, \gls{cbmlp} and convolutional network. Furthermore, we propose a novel technique of neural weight stepping, where subsequent frames of a video are encoded as low-entropy parameter updates. To assess the feasibility of the considered approaches, we will test the video compression performance on several high-resolution video datasets and compare against existing conventional and neural compression techniques.
\end{abstract}

\begin{figure}[!b]
    \centering
    \includegraphics[width=\textwidth]{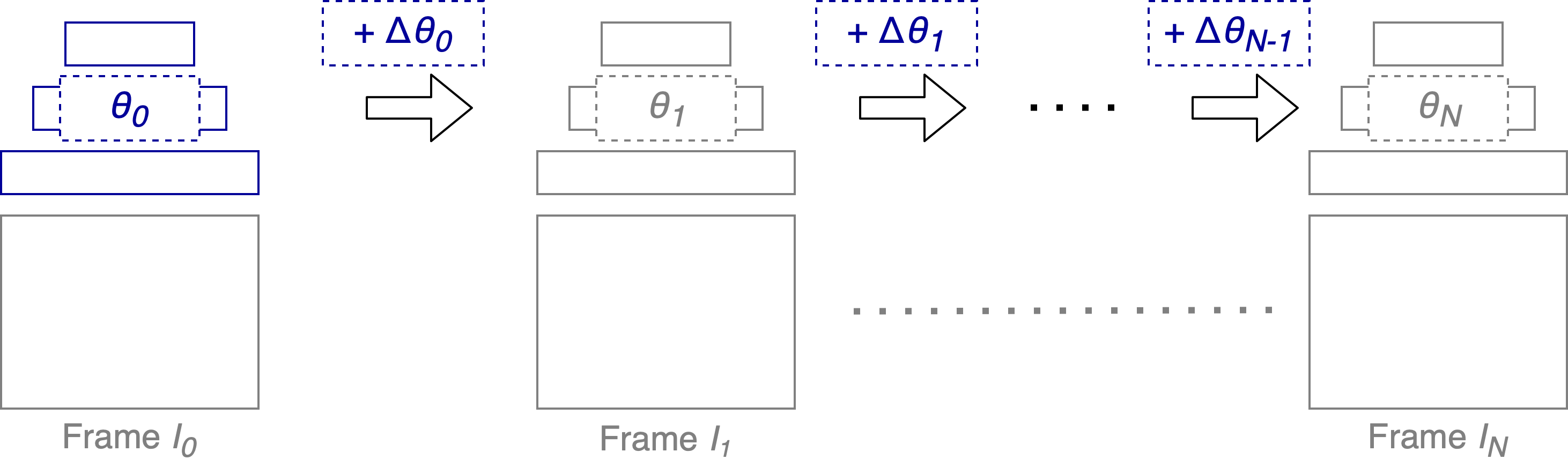}
    
    \caption{The weight step video compression technique employs a neural network to parameterize a single image of the sequence and encodes subsequent frames as low-entropy weight steps applied to the network. Only parameters in dark blue need to be transmitted.}
    \label{fig:intro_diagram}
\end{figure}

\section{Introduction}
    Video compression is a relevant topic of ongoing research in the current stage of the digital age. The amount of shared video data continues to grow due to the increasing popularity of services such as video-on-demand, real-time communication, or distance learning~\cite{Ding2021}. To accommodate for this rapid expansion with the available bitrate budget, efforts to maximize the amount of recoverable information for every physically transmitted bit play a crucial role. This optimization process relies primarily on various compression techniques. These techniques can be divided into conventional, deterministic video coding methods~\cite{Tudor1995, Sikora1997, h264, h265}, widely employed in practice, as well as fairly new neural video compression techniques~\cite{Chen2018, Dong2020, Ma2020, Rippel2019, Liu2020, Lu2019, Wu2018, Djelouah2019}.

    Most often, the video compression techniques based on neural networks exhibit close resemblance to the traditional pipelines, that is, they train an encoding module to produce a compressed bitstream that a decoding module can reconstruct the original message from~\cite{Chen2018, Dong2020, Ma2020, Rippel2019, Liu2020, Lu2019, Wu2018, Djelouah2019}. We propose an alternative solution, where a neural network is used directly as the compressed representation of the video. Furthermore, we propose a technique, where the shared information in consecutive frames is exploited. This is done by encoding an image sequence $I_0, I_1, ..., I_N$ as low-entropy network weight updates $\Delta \theta_1, ..., \Delta \theta_{N-1}$ applied to the network parameters $\theta_0$ representing the anchor frame $I_0$, as demonstrated in~Figure~\ref{fig:intro_diagram}.

    If successful, this approach will have the considerable advantage of being data-agnostic, unlike other neural video compression techniques published to date that must be pretrained on an external dataset~\cite{Chen2018, Dong2020, Ma2020, Rippel2019, Liu2020, Lu2019, Wu2018, Djelouah2019}. How the proposed method compares to the state-of-the-art video codecs~\cite{h264, h265} remains to be tested in the experimental stage.

\section{Related Work}
    Several published works provide some foundation components of the solution proposed in this work. The technique of parameterizing images using neural network weights (either coordinate- or convolution-based) has already been explored to some degree~\cite{Dupont2021, Ulyanov2020}. Further, neural networks have already been employed for compression, albeit in fundamentally different schemes. Here, we provide a summary of these past contributions.

    \subsection{Neural Implicit Representations}
        The proposed techniques are closely related to the recent work on compression with \glsfirst{cbmlp} networks~\cite{Dupont2021}. There, natural images are encoded using an \gls{mlp} network that maps input space coordinates to the appropriate pixel color values. The experiments demonstrate that COIN is \textbf{not} capable of beating the state-of-the-art compression schemes (including standardized codecs and neural compression baselines)~\cite{Dupont2021}. However, the potential for applying neural implicit compression for videos has not been explored. Furthermore, the proposed parameter step technique can potentially enhance the performance compared to methods based on direct frame encoding.

        Furthermore, a surge in interest pertaining to 3D representation learning was sparked by the introduction of the \gls{nerf} technique~\cite{Mildenhall2020}. Similarly to COIN, the approach heavily relies on a \gls{cbmlp}, albeit with an input space of higher dimensionality; three spatial coordinates ($x,y,z$) and two viewing direction coordinates ($\theta, \phi$). These additional dimensions are learned in the same manner as the temporal dimension of a video representation would be. This approach, in fact, will be one of the configurations tested against the weight step technique proposed in this work.

    \subsection{Deep Convolutional Priors}
        A growing body of work addresses the potential of encoding images in parameters of a convolutional neural network. This technique has been demonstrated to be useful for several image inverse problems, such as inpainting, super-resolution, or denoising~\cite{Ulyanov2020}. Some work on extending these techniques for similar inverse tasks in video domain has been done~\cite{Lei2020}, however, not in the context of image compression. The prior introduced by the use of convolutional kernels appears to have a capability of biasing the solution towards a domain of narrow texture distributions, based on the existing information in the synthesized signal. We hypothesize that this domain narrowing effect can be beneficial for finding compressed representations of the signal.

    \subsection{Neural Video Compression}
        The approach of encoding the signal into a compressed transmittable representation and then reconstructing back to the original representation~\cite{Balle2017, Chen2018, Dong2020, Ma2020} is known as end-to-end neural video coding~\cite{Ding2021} (other variants identified in~\cite{Ding2021} include up-sampling approaches and modularized schemes that are combined with traditional coding frameworks). We focus on the end-to-end coding variant since it is most closely related to the idea proposed in this work. Recently explored neural image compression techniques tend to rely on the training of a convolutional autoencoder network that reconstructs the output from the compressed bitstream~\cite{Balle2018, Minnen2018,Lee2019}. Analogous methods have been tested in the video domain~\cite{Rippel2019, Liu2020, Lu2019}. Further approaches to neural video coding take advantage of the correlations in the temporal domain~\cite{Wu2018, Djelouah2019}. However, all cited neural video compression techniques require a network to be pretrained on an external dataset before use, which carries a significant computational cost and a risk of overfitting. Our approach is data-agnostic in the sense that it requires no pretraining on any external datasets and adapts to any new video.

\section{Compressing Video as Parameter Steps}

    We wish to learn whether video compression can be achieved by relying on weight changes applied to a neural network that encodes a two-dimensional image representation of an anchor frame (for example. the first frame in the sequence). This approach could have several benefits. First, directly encoding a single anchor frame in the weights of a neural network should be less computationally demanding than encoding a three-dimensional video representation; for architectures based on convolutions the model size tends to grow exponentially with every new dimension of the convolution. If representational neural networks implicitly learn a set of visual abstractions about the image, then making small changes to the network weights may allow to reuse these abstractions and therefore, be an efficient way of encoding motion pictures.
    
    There are at least two fundamentally different techniques for representing images as a neural network. An image can be reconstructed as i) the response of a \textbf{convolutional network} to a fixed random input vector~\cite{Ulyanov2020}, or ii) the output distribution of an \textbf{\gls{cbmlp}} network to a set of coordinate inputs (one for each pixel)~\cite{Sitzmann2020, Tancik2020}.

    \subsection{Convolutional Network}
        Methods of encoding images in the weights of convolutional neural networks have been proposed in seminal works such as~\cite{Ulyanov2020}. The authors claim that the sole use of convolutional architectures is a source of useful prior for natural images. The setting involves a convolutional network with parameters $\theta$ that maps a random fixed noise input $z$ to a given output approximation $\hat{I}$ with a function $f$. During training, the error between the approximation $\hat{I}$ and ground truth image $I$ is minimized.
    
    \begin{equation}
        f(\theta, z) = \hat{I}
    \end{equation}

    \subsection{Coordinate-based MLP Network}
        As in COIN~\cite{Dupont2021}, we can encode a signal in any number of output dimensions, whether it is two spatial dimensions for a single anchor frame, or a video sequence with one more dimension in the output distribution. This setting is largely based on the recent developments~\cite{Sitzmann2020, Tancik2020} related to encoding an image as the output distribution $f$ of a \gls{mlp} that maps from coordinates ($x,y$) to pixel color values $\hat{c}(x,y)$ at the queried coordinate.
    
        \begin{equation}
            f(\theta, x, y) = \hat{c}(x,y)
        \end{equation}
    
        For the weight step technique, we only need to encode a single two-dimensional image $I$ containing the anchor frame. The approximated image $\hat{I}$ is obtained by aggregating predicted color $\hat{c}(x,y)$ for all coordinates of ground-truth image $I$.
        
        Furthermore, the capability of \gls{cbmlp} networks to produce high-frequency detail in the output distribution has only been achieved recently by utilising techniques such as random Fourier feature encoding~\cite{Tancik2020}, or sinusoidal network activations~\cite{Sitzmann2020}. In this work, we will test both approaches in the first stage of experiments.
        
        The optimization process for both \gls{cbmlp} and convolutional variants is similar, where the reconstruction error between the ground truth image $I$ and the approximation $\hat{I}$ resulting from $f_{(\theta)}$ is minimized. For clarity, we describe the high-level image approximation function of each variant as $f_{(\theta)}$ from now on, abstracting away the differences such as the need for a random input vector $z$, or aggregation of all input coordinates $(x,y)$.

    \subsection{Weight Step Encoding}
        For both representation paradigms, the setting is similar; we want to first encode the parameters $\theta_0$ that result in the generation of the first image $I_0$ of the video sequence. Any given neural network topology determines a function $f_{(\theta)}$ that yields a specific 2D output.
    
        \begin{equation}
            f_{(\theta_0)} = I_0
        \end{equation}
        
        Then, at each temporal step $t$, we wish to find a low entropy parameter shift $\Delta \theta_t$ that results in the image at $I_{t+1}$.
        
        \begin{equation}
            f_{(\theta_t + \Delta \theta_t)} = I_{t+1}
        \end{equation}
        
        
        This, however, should be subject to constraints that enforce efficient encoding. Rather than computing any parameter shift resulting in the subsequent image, we want to find one such that its entropy is low. This can be done by enforcing sparsity in the weight step representation, similarly to existing techniques for network sparsity~\cite{Louizos2018, Qiao2020, LeCun1990, Molchanov2017}. A promising approach would be to add a regularization term $\mathcal{R}(\Delta\theta_t)$ to the objective function. As analyzed in~\cite{Louizos2018}, regularizing the weight parameters with $\ell_0$ norm is not practical, since the the differentiability condition is not satisfied. To that end, we utilize the technique proposed by the same authors in~\cite{Louizos2018} based on the introduction of non-binary stochastic gates and a surrogate sparsity objective. Further, we want to explore two techniques of sparsity regularization. The first technique will involve maximizing sparsity of the direct parameter update array. This will reward solutions where only a small number of network weights is updated at each step:
        
        \begin{equation}
            \mathcal{R}(\Delta\theta_t) = ||\Delta\theta||_0 
        \end{equation}
        
        Alternatively, using the second technique, sparsity can be achieved in a domain orthogonal to the spatial domain. In this case, we propose to test a technique where sparsity is enforced in the spatial frequency domain instead. This can be done by applying the same sparsification technique to the \gls{dct} of each layer block (for convolutional network, \gls{dct} is applied to individual sequences of aligned kernel elements), similarly to~\cite{Cheinski2020DCTConvCF}. The technique allows the parameter update solution to involve changes to a large number of network weights while preserving low information content.
        
        \begin{equation}
            \mathcal{R}(\Delta\theta_t) = ||\textrm{DCT}(\Delta\theta)||_0 
        \end{equation}
        
        Finally, the optimization process itself can be done in at least two different ways. The network can be trained in two stages, where the first stage optimizes only the starting point $\theta_0$ while the second stage optimizes an $(N-1)$-long sequence of $\Delta \theta_t$ originating from that point. Another way would be to train this sequence in a single-stage process, where the complete sequence $\left\{(\theta_0),(\theta_0 + \Delta \theta_0), ...,(\theta_0 + \sum_{\substack{t=0}}^{N-1} \Delta\theta_t) \right\}$ is to be found.

\section{Experimental Protocol}
    
    We plan to test on several video datasets. Our primary focus will be on high-resolution datasets such as UVG (1080p and 4K)~\cite{uvg-dset} and MCL-JVC (1080p)~\cite{mcl-jcv-dset}. Furthermore, we will run additional experiments on a lower-fidelity dataset of Video Trace Library (288p)~\cite{yuv-dset}.
    
    \subsection{Base Architectures}
    \label{sec:base_arch}
    
        Our two basic architecture types are \gls{cbmlp} and a convolutional neural network. During the experiments, we will want to compare several alterations of each model corresponding to a different parameter count.

        Furthermore, we will want to test how the weight step compression techniques compare to equivalent models that directly encode the entire video, with no weight stepping applied. For the \gls{cbmlp}, the full video can be captured by learning one additional input dimension $t$ for time ($f(\theta, x, y) = \hat{c}(x,y)$ becomes $f(\theta, x, y,t) = \hat{c}(x,y,t)$). For the convolutional module, one solution to approach this is to define a fixed sequence of input tensors $\{z_0, z_1, ..., z_N\}$, one $z_i$ for each of $N$ video frames. Similarly, to a single $z$, this input activation is predefined before use and remains the same for any encoded video.

    \subsection{Architectural Ablations}
    
        Before evaluating the core results, we will want to run several design-related experiments for our method.

        Both the SIREN approach~\cite{Sitzmann2020} and random Fourier feature encoding~\cite{Tancik2020} appear to be a valid design choice for coordinate-based \gls{mlp} variant. However, it is not clear how the two approaches compare in the context of end-to-end compression. To better understand, which approach can encode signals in a more parameter-efficient manner, we compress each image in the three datasets using the two approaches with the same set of underlying layer topology.

        For the convolution-based approach, we would like to test whether the application of long-distance skip connections (like in some of the configurations proposed in the Deep Image Prior work~\cite{Ulyanov2020}) impacts the reconstruction quality. We hypothesize that the unimpeded flow of information through the skip connections can benefit the network's capability to reuse encoded abstractions.
        
    \subsection{Model Capacity Tests}
        Several variables are expected to greatly influence the resulting distortion and bitrate. Here, we focus on the model size, but for longer videos, it may be beneficial to split the complete sequence into slices with fewer frames and repeat the process independently for each. This can not only affect the quality of reconstruction but also encoding time (the time it takes to encode a complete video). For that reason, we will test several model sizes and their capability to encode video sequences of various length, as well as the time it takes to encode a complete video. Then, the configurations that offer the best distortion-to-bitrate trade-off will be selected for the subsequent core evaluation stage.

    \subsection{Core Evaluation}
        The baselines will include common video compression codecs such as H.264~\cite{h264} and H.265~\cite{h265} as well as two recent methods of neural video compression of DVC~\cite{Lu2019} and VCII~\cite{Wu2018}. For the latter, the models pretrained by the authors that are publicly available will be employed. We will compare these baselines against the best performing variants of our method resulting from the preceding experiments described above. This leads to a total of 10 tested compression techniques, as listed in Table~\ref{tab:core_results}.
        
        \begin{table}[!h]
            \centering
            \caption{Set of Tested Compression Techniques}
            \begin{tabular}{l l}
                \toprule
                Type & Variant \\
                \midrule
                 & Full Sequence (as described in \ref{sec:base_arch})\\
                CNN & Weight Stepped - Two Stage\\
                 & Weight Stepped - Single Stage\\
                 \midrule
                 & Full Sequence (as described in \ref{sec:base_arch})\\
                CbMLP & Weight Stepped - Two Stage\\
                 & Weight Stepped - Single Stage\\
                 \midrule
                Video Codecs & H.264~\cite{h264}\\
                & H.265~\cite{h265}\\
                 \midrule
                Neural Video Compression & DVC~\cite{Lu2019}\\
                & VCII~\cite{Wu2018}\\
                \bottomrule
            \end{tabular}
            \label{tab:core_results}
        \end{table}
        
        Our results will contain rate-distortion plots for each tested method and dataset. Furthermore, we will provide further analysis of total encoding and decoding time depending on the video length for each method since these can be expected to vary significantly.

\section{Conclusion}

    Compression using neural implicit representations has only been tested for 2D images, where a degree of compressive capability has been demonstrated. In this work, we would like to explore the potential of similar approaches to video compression. Furthermore, a previously unexplored technique of low-entropy weight steps is described, which aims to learn network state sequences with high shared information content.


\end{document}